\documentclass{article}

\usepackage{PRIMEarxiv}

\usepackage[utf8]{inputenc} 
\usepackage[T1]{fontenc}    
\usepackage{hyperref}       
\usepackage{url}            
\usepackage{booktabs}       
\usepackage{amsfonts}       
\usepackage{nicefrac}       
\usepackage{microtype}      
\usepackage{lipsum}
\usepackage{fancyhdr}       
\usepackage{graphicx}       
\graphicspath{{media/}}     

\usepackage{cite}
\usepackage{amsmath,amssymb,amsfonts}
\usepackage{algorithmic}
\usepackage{graphicx}
\usepackage{textcomp}
\usepackage{xcolor}
\usepackage{multirow}
\usepackage{booktabs}
\usepackage{makecell}
\usepackage{arydshln}
\usepackage{geometry}
\usepackage{algorithm,algorithmic}
\usepackage{url}
\usepackage{multicol}

\usepackage{caption}
\hypersetup{
    colorlinks=true,
    linkcolor=black,
    filecolor=black,
    urlcolor=black,
    citecolor=black
}

\title{A Fully Interpretable Statistical Approach for Roadside LiDAR Background Subtraction
\thanks{\textit{\underline{Citation}}: 
A. Iglesias et al., \textit{"A Fully Interpretable Statistical Approach for Roadside LiDAR Background Subtraction,"} \textbf{The IEEE International Conference on Vehicular Electronics and Safety (ICVES2025)}, DOI: \textit{\href{https://doi.org/10.1109/ICVES65691.2025.11376398}{10.1109/ICVES65691.2025.11376398}}.}
}

\author{
Aitor Iglesias\textsuperscript{1,2},
Nerea Aranjuelo\textsuperscript{1},
Patricia Javierre\textsuperscript{3},
Ainhoa Menendez\textsuperscript{3},\\
\textbf{Ignacio Arganda-Carreras\textsuperscript{2,4,5,6},
Marcos Nieto\textsuperscript{1}}
\vspace{1em}\\
\small
\textsuperscript{1}Fundación Vicomtech, Basque Research and Technology Alliance (BRTA), Donostia-San Sebastián, Spain\\
\textsuperscript{2}University of the Basque Country (UPV/EHU), Donostia - San Sebastián, Spain\\
\textsuperscript{3}CAFSignalling, Amorebieta, Spain\\
\textsuperscript{4}IKERBASQUE, Basque Foundation for Science, Bilbao, Spain\\
\textsuperscript{5}Donostia International Physics Center (DIPC), Donostia - San Sebastián, Spain\\
\textsuperscript{6}Biofisika Institute, Leioa, Spain
}

\fancypagestyle{firstpage}{
  \fancyhf{} 
  \fancyfoot[C]{\footnotesize \textcopyright~2025 IEEE. This is the author’s version of the paper published in The IEEE International Conference on Vehicular Electronics and Safety (ICVES2025). The final published version is available at \href{https://doi.org/10.1109/ICVES65691.2025.11376398}{https://doi.org/10.1109/ICVES65691.2025.11376398}.}
}

\begin{document}
\maketitle
\thispagestyle{firstpage}

\begin{abstract}
We present a fully interpretable and flexible statistical method for background subtraction in roadside LiDAR data, aimed at enhancing infrastructure-based perception in automated driving. Our approach introduces both a Gaussian distribution grid (GDG), which models the spatial statistics of the background using background-only scans, and a filtering algorithm that uses this representation to classify LiDAR points as foreground or background. The method supports diverse LiDAR types, including multiline 360\textdegree and micro-electro-mechanical systems (MEMS) sensors, and adapts to various configurations. Evaluated on the publicly available RCooper dataset, it outperforms state-of-the-art techniques in accuracy and flexibility, even with minimal background data. Its efficient implementation ensures reliable performance on low-resource hardware, enabling scalable real-world deployment.
\end{abstract}

\keywords{Automated Driving \and Interpretability \and Point Cloud \and Object Detection}

\begin{multicols}{2}

\section{Introduction}

The integration of perception functions into infrastructure is essential for advancing automated driving (AD) systems. When infrastructure elements such as traffic lights, road signs, or smart sensors embedded in roadways detect objects, they can significantly enhance a vehicle's ability to perceive its environment. This infrastructure can interact with connected vehicles, corroborating the vehicle’s sensor data or providing additional information, which is particularly valuable when a vehicle’s sensors are obstructed or malfunctioning. By supplementing vehicle sensors with real-time data from infrastructure, the overall safety, reliability, and effectiveness of AD systems can be improved, contributing to smoother, more secure driving experiences~\cite{lu2014connected}.

In this context, LiDAR offers significant advantages over traditional camera-based systems in autonomous driving. While cameras capture two-dimensional images, LiDAR provides precise three-dimensional mapping of the environment, enabling vehicles to measure distances to objects with high accuracy. This enhanced spatial awareness allows for better navigation and obstacle detection, particularly in complex or dynamic environments. Although cameras are useful for visual recognition, the depth perception provided by LiDAR adds a layer of safety and robustness to autonomous systems, making it a preferred sensor for AD systems.

Equally important is the interpretability of the algorithms that drive these systems. In autonomous driving, it is critical that the decision-making processes of algorithms be transparent and understandable. Interpretability ensures that engineers, regulators, and users can comprehend how and why certain decisions are made, fostering accountability and trust in the system. Moreover, an interpretable algorithm facilitates the identification and correction of potential flaws or biases, thereby improving system performance. For the responsible deployment of AD systems, particularly in safety-critical applications such as autonomous driving, interpretability is not only beneficial but a crucial aspect of ensuring alignment with human values, legal standards, and regulatory requirements.

The main contributions of this work can be summarized as follows:
\begin{itemize}
    \item We propose a fully interpretable approach for roadside LiDAR background subtraction.
    \item Our method outperforms existing state-of-the-art techniques in terms of accuracy and robustness.
    \item The proposed approach is highly flexible and adaptable to various LiDAR configurations, including different LiDAR models and both single- and multi-sensor setups.
    \item We evaluate our method on a public dataset, promoting reproducibility and enabling direct comparison with future approaches.
\end{itemize}

\section{State of the art}

Background subtraction is an initial and important step in processing roadside LiDAR data. The background refers to the static scenario captured by the sensor. Background points can cause serious interferences to target detection and classification. Filtering the background is crucial because the target objects, such as vehicles and pedestrians, often constitute only a small percentage of the total data points collected by roadside LiDAR. By excluding irrelevant background points (e.g., buildings, trees, ground surface, road facilities), the computational load of subsequent data processing (like clustering, object identification, and tracking) can be significantly reduced, and the accuracy of vehicle and pedestrian recognition can be improved.

\subsection{Roadside LiDAR datasets}

Recent years have seen an increasing number of roadside LiDAR datasets being released to support research in infrastructure-based perception. These datasets vary significantly in terms of sensor configuration, scene coverage, and annotation quality.

Some of the earliest datasets, such as BAAI-VANJEE~\cite{BAAI-VANJEE}, provide data from a single roadside LiDAR unit. These setups are limited in spatial coverage but offer annotated scans for evaluating basic perception tasks under realistic urban conditions.

More recent efforts have included datasets with multiple scenarios and sparse sensor deployments. For instance, DAIR-V2X-I~\cite{DAIR-V2X-I} and Rope3D~\cite{Rope3D} provide annotated point clouds from different locations, enabling more diverse background structures, but are still restricted in terms of sensor variety and synchronization complexity.

A different approach is seen in datasets like LUMPI~\cite{LUMPI}, which focus on a single, complex intersection and employ a dense setup with up to five rotating LiDARs. This configuration enables accurate multi-perspective labeling and tracking, offering high-quality data for evaluating multi-view background modeling strategies.

Finally, datasets such as RCooper~\cite{RCooper} incorporate multiple types of LiDAR sensors (e.g., micro-electro-mechanical systems --MEMS-- and rotating LiDARs) across different infrastructures. This heterogeneity introduces new challenges for background subtraction, including cross-device calibration, varying point densities, and dynamic occlusion patterns, making them particularly relevant for research on generalizable filtering methods.

\begin{figure*}[!ht]
    \centering
    \includegraphics[width=\textwidth]{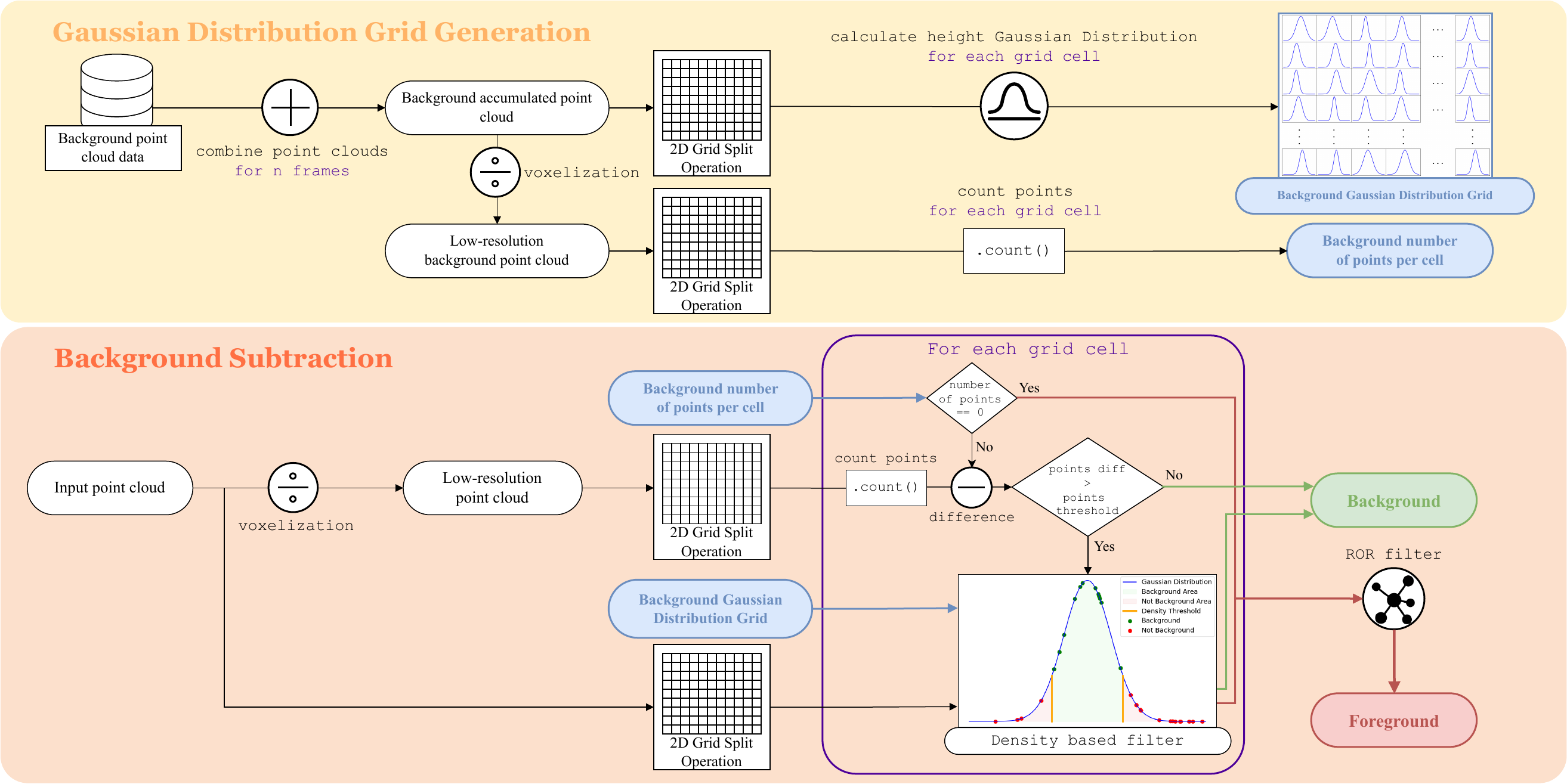}
    \caption{Diagram of our roadside LiDAR background subtraction algorithm. From top to bottom: (1) \textit{Gaussian Distribution Grid generation}, which creates a statistical representation of the background using Gaussian distributions; this representation serves as the foundation for the subsequent background subtraction process. (2) \textit{Background Subtraction}, which uses the learned Gaussian distributions to classify each LiDAR point as either background or foreground.}
    \label{fig:methodology}
    \vspace{-10pt}
\end{figure*}

\subsection{Perception for roadside LiDAR}

A range of strategies has been proposed for background subtraction in roadside LiDAR, each leveraging different aspects of point cloud data and varying in their adaptability to different sensor types and deployment conditions.

Many of the strategies rely on spatial representations to capture the geometric regularities of the environment. These methods commonly divide the scene into grids, voxels, or polar coordinates to support background modeling. Voxel-based techniques, such as 3D density statistic filtering (3D-DSF), estimate typical point densities over time to distinguish dynamic from static regions~\cite{17BinLv,21WangGuojun}. Other approaches make use of rasterized grids, azimuth angles, channel IDs, or intensity values to construct background profiles. For instance, azimuth-based filtering leverages the fixed scanning pattern of rotating LiDARs to model height distributions per angle~\cite{19JunxuanZhao,22LuyangWang}, while channel- and intensity-based methods exploit laser-specific information to separate background structures~\cite{19JianqingWu,19JianqingWu2,22HuiLiu}. However, these representation-driven approaches often suffer from limited flexibility: azimuth-based methods are incompatible with non-rotational LiDAR architectures like MEMS or Risley Prism, and intensity- or channel-based techniques require device-specific calibration, as different sensors return inconsistent values for the same object~\cite{22TianyaZhang,22YingjiXia}.

In parallel, more traditional methods rely on background modeling through temporal comparison. Frame-difference techniques construct a background model—commonly via mean background modeling (MBM)—and detect foreground points as deviations from this baseline, often using 2.5D grid representations that incorporate elevation~\cite{18JianyingZheng,22JianqingWu}. Reference-based methods, on the other hand, compare current scans to one or more preselected ''clean'' background scans, assuming that stable points across time belong to the background~\cite{20YanjieSong,20YuepengCui}. While these approaches are conceptually simple and computationally efficient, they can be vulnerable in crowded urban environments where obtaining reliable background scans is challenging, and where scene dynamics reduce the reliability of static models.

Finally, a limited but growing number of deep learning methods have emerged. These techniques aim to improve adaptability and accuracy by learning background features from data classifying points directly through neural networks~\cite{23ShanglianZhou,23ZirenGong}. While promising, they typically require large annotated datasets and still face challenges in generalizing across different LiDAR sensors and deployment scenarios.

However, it is important to note that none of the aforementioned methods have been evaluated using publicly available roadside LiDAR datasets. This lack of standardized benchmarks hinders direct comparison between approaches and limits reproducibility, making it difficult to assess the generalizability and robustness of current solutions.


\section{Methodology}

To address the challenge of background subtraction in roadside LiDAR data, we developed a novel algorithm designed with interpretability in mind. According to the definition of interpretable design proposed by Naranjo et al.~\cite{xai_ontology}, our method qualifies as interpretable because its inner workings are inherently transparent: its outputs can be understood directly from the model’s structure and operation, without requiring post hoc explanation. Beyond being interpretable, the algorithm is also highly flexible. Unlike state-of-the-art methods that primarily target rotating LiDAR sensors, our approach supports a broader range of devices, including emerging non-rotating technologies such as MEMS and Risley Prism LiDARs, which have recently gained popularity. Moreover, our method can operate with either a single LiDAR sensor or multiple sensors, enhancing spatial coverage and robustness at the cost of increased computational demand.

The proposed method has two phases as illustrated in Figure~\ref{fig:methodology}: generating a \textbf{Gaussian Distribution Grid} (GDG) from background-only scans, and a subsequent algorithm that uses this GDG to classify points as background or foreground 

\subsection{Gaussian Distribution Grid generation}

The first phase of the method focuses on building a robust and homogeneous background representation by leveraging two types of point clouds: an accumulated point cloud and a low-resolution voxelized version.

To generate the accumulated point cloud, multiple point cloud scans containing only background elements are combined. This accumulation captures a richer and more complete representation of the background, including dynamic elements such as moving tree branches, and helps mitigate the impact of sensor vibrations or noise. From this representation, a low-resolution point cloud is created through a voxelization process. Voxelization contributes to the uniformity of the representation, improving robustness to sensor irregularities.

Both the accumulated (rich) point cloud and the voxelized (low-resolution) point cloud are then divided into a 2D grid based on the $(x, y)$ coordinates of the points. The voxelized cloud is used to count the number of points within each grid cell, providing a compact and stable statistical descriptor of point density. Meanwhile, the richer accumulated cloud is used to estimate a Gaussian distribution of the height (z-value) of the points within each grid cell.

For this first phase, two hyperparameters are required additionally to the input point clouds:
\begin{itemize}
    \item $\texttt{voxel\_size}$: Size of the voxels (3D array) for the generation of the low-resolution representation of the point cloud.
    \item $\texttt{cell\_size}$: Size of the cells (2D array) of the Gaussian Distribution Grid. $\texttt{cell\_size}$ must be bigger than $\texttt{voxel\_size}$.
\end{itemize}

This dual representation is used as an input of the next phase of the algorithm. Using this data enables the method to both capture local spatial variations and model the statistical distribution of the background in each region. The Gaussian distributions are essential for modeling the expected height values and their variances, allowing the background subtraction algorithm to adapt effectively to local terrain variations and to distinguish between background and foreground in subsequent processing stages.

\subsection{Background subtraction}

The second phase of the method performs background subtraction using the GDG generated in the previous step. The algorithm for this phase is described in Algorithm~\ref{alg:background_subtraction}. The GDG is represented by the variable $\texttt{gdg}$, where each cell contains statistical information: the mean $\mu$, standard deviation $\sigma$, the maximum value of the Gaussian distribution $\texttt{max\_density}$, and the number of points $\texttt{num\_points}$.

In addition to the input point cloud, this phase requires the following hyperparameters:

\begin{itemize}
    \item $\texttt{voxel\_size}$: Size of the 3D voxels used to generate the low-resolution version of the point cloud. Must be the same as used in the GDG generation.
    \item $\texttt{cell\_size}$: Size of the 2D grid cells in the GDG. Must also match the value used in the previous phase.
    \item $\texttt{th\_points}$: Threshold for the difference in points between the current scan and background model for a point to be considered a foreground candidate.
    \item $\texttt{th\_density}$: Minimum percentage of the cell's maximum Gaussian density for a point to be classified as background.
    \item $\texttt{neighbors}$: Number of neighbors considered in the Radius Outlier Removal (ROR) step.
    \item $\texttt{radius}$: Search radius used in the ROR step.
\end{itemize}

In the algorithm, the input point cloud is denoted as $\texttt{input\_pc}$. Background and foreground points are stored in the variables $\texttt{bg\_pts}$ and $\texttt{obj\_pts}$, respectively.

\begin{center}
\footnotesize
\noindent\rule{0.96\columnwidth}{0.8pt}
\captionof{algorithm}{Background Subtraction}
\noindent\rule{0.96\columnwidth}{0.8pt}
\vspace*{-8pt}
\label{alg:background_subtraction}
\begin{algorithmic}
\STATE
\COMMENT{Voxelization}

\STATE $\texttt{low\_res\_pc} \gets \texttt{voxelize}(\texttt{input\_pc}, \texttt{voxel\_size})$
\STATE
\item[]

\COMMENT{Point counting}

\FOR{each \texttt{pt} in $\texttt{low\_res\_pc}$}
    \STATE $cell \gets \texttt{get\_cell\_id}(pt.x, pt.y, \texttt{cell\_size})$
    \STATE $\texttt{point\_count}[cell] \gets \texttt{point\_count}[cell] + 1$
\ENDFOR
\STATE
\item[]

\COMMENT{Background Subtraction}

\FOR{each \texttt{pt} in $\texttt{input\_pc}$}
    \STATE $cell \gets \texttt{get\_cell\_id}(\texttt{pt.x}, \texttt{pt.y}, \texttt{cell\_size})$
    \IF{$\texttt{gdg}[cell].\texttt{num\_points} == 0$}
        \STATE $bg \gets \texttt{false}$
    \ELSIF{$\texttt{point\_count}[cell] > (\texttt{gdg}[cell].\texttt{num\_points} + \texttt{th\_points})$}
        \STATE $\mu \gets \texttt{gdg}[cell].\texttt{mu}$
        \STATE $\sigma \gets \texttt{gdg}[cell].\texttt{sigma}$
        \STATE $\texttt{max\_dens} \gets \texttt{gdg}[cell].\texttt{max\_density}$
        \STATE $\texttt{dens} \gets \texttt{normalPDF}(\texttt{pt.z}, \mu, \sigma)$
        \STATE $bg \gets (\texttt{dens} > \texttt{max\_dens} \times \texttt{th\_density})$
    \ELSE
        \STATE $bg \gets \texttt{true}$
    \ENDIF

    \IF{$bg$}
        \STATE $\texttt{bg\_pts}.\texttt{push\_back}(\texttt{pt})$
    \ELSE
        \STATE $\texttt{obj\_pts}.\texttt{push\_back}(\texttt{pt})$
    \ENDIF
\ENDFOR
\STATE
\item[]

\COMMENT{Radius Outlier Removal}

\STATE $\texttt{obj\_pts} \gets \texttt{radius\_outlier\_removal}(\texttt{obj\_pts},$\\
\STATE \hspace{5em} $\texttt{radius}, \texttt{neighbors})$
\noindent\rule{0.96\columnwidth}{0.8pt}
\end{algorithmic}
\vspace{-11pt}
\end{center}

The algorithm is structured into four main stages: voxelization, point counting, background subtraction, and Radius Outlier Removal (ROR).

Initially, a low-resolution version of the input point cloud is generated using voxelization. Then, both the original and voxelized point clouds are divided into grid cells using the same parameters as the background model. The number of points per cell in the low-resolution cloud is computed and compared with the corresponding counts in the GDG.

The classification of each point proceeds as follows:

\begin{itemize}
    \item If the background model has no points in a given cell, all points in that cell are classified as foreground.
    \item If the difference in point count between the scan and the model is below the threshold $\texttt{th\_points}$, all points are classified as background.
    \item If the difference exceeds the threshold, each point’s height is evaluated against the Gaussian distribution of the cell. If the computed probability density is below $\texttt{th\_density} \times \texttt{max\_density}$, the point is classified as foreground.
\end{itemize}

Finally, a ROR filter is applied to the set of foreground points to eliminate isolated noise that may result from sensor vibrations or measurement errors. The ROR filter works by analyzing the spatial neighborhood of each point: a point is considered an outlier, and is consequently removed if it has fewer than a specified number of neighboring points within a given radius. This process ensures that only spatially consistent clusters of points ---likely to represent real objects--- are retained, while isolated or spurious points are discarded.

\section{Experiments}

To assess the performance of the proposed algorithm, we conducted a series of experiments to evaluate its flexibility across different LiDAR configurations, as well as its accuracy and efficiency. This section provides details about the experimental setup, the datasets used, the evaluation metrics, and comparisons with existing methods.

\subsection{Dataset}

To validate our method, we used the publicly available RCooper dataset~\cite{RCooper}. We followed the train and validation split proposed in the original article. To obtain the background scans, we extracted the points corresponding to dynamic objects from the training sample. As mentioned in the Methodology section, our method does not require training; it only requires a few background scans as a reference.

We conducted experiments using different numbers of background scans to analyze the impact of this parameter. Specifically, we tested 10 scans (1 second), 25 scans (2.5 seconds), 50 scans (5 seconds), and 100 scans (10 seconds), randomly selecting scans from the training sample.

RCooper provides two different scenarios for testing our method: \textit{corridor}, which includes two multiline 360-degree LiDAR sensors, and \textit{intersection}, which includes two multiline 360-degree LiDAR sensors and two MEMS LiDAR sensors. The point clouds of the dataset contain data from all sensors combined. However, we aimed to evaluate the performance of our algorithm using individual sensors as well as sensor combinations. Additionally, we aimed to assess performance across different types of LiDAR sensors.

To achieve this, we prepared validation samples based on the original validation set. For the corridor scenario, we created two validation samples: one containing individual LiDAR data and another with all point clouds combined. For the intersection scenario, we generated more sequences due to the greater variety of data. First, we separated the sensors individually, dividing the validation sample into point clouds from multiline LiDAR and MEMS LiDAR. For the combined point clouds, we defined three validation samples: one with only multiline LiDAR sensors, another with only MEMS LiDAR sensors, and a third combining all sensors.

The labels of the dataset are: \textit{car}, \textit{pedestrian}, \textit{truck}, \textit{bus}, \textit{bicycle}, \textit{motorcycle}, \textit{tricycle}, \textit{construction}, \textit{huge vehicle} and \textit{signal}. For our experiments we exclude the \textit{signal} class since it is considered a part of the background. Also, the validation set of the corridor sequence does not contain samples of the \textit{huge vehicle} class.

This sampling approach allowed us not only to determine which scenario our method performs best in, but also to assess which sensor is most suitable for our method and whether combining sensor data improves performance.


\subsection{Evaluation}

To ensure a fair and consistent evaluation across all experiments in this section, we first conducted a preliminary search to identify effective hyperparameter values. Unless otherwise stated, the following configuration was found to yield the best empirical results: a voxel size of $0.1$ along all axes, a cell size of $0.2$ for both the x and y axes, a points threshold of $2$, a density threshold of $0.3$, and $4$ neighbors and a radius of $0.8$ for the Radius Outlier Removal step.

For the experiments, we first evaluate the performance of the background subtraction at the point level. To do so, we measure precision, recall, F1 score, and Intersection over Union (IoU) of the background points. The results for these metrics are shown in Table~\ref{tab:iou}






The ablation study reveals notable differences in performance between the two evaluated scenarios and sensor configurations. The intersection scenes consistently outperform the corridor setting across all metrics, particularly in terms of IoU, which is the most indicative measure of segmentation quality. This suggests that the corridor scenes present greater challenges for the proposed algorithm, likely due to more constrained viewpoints or reduced scene variability, whereas the intersection scenario benefits from richer spatial information and more diverse perspectives.

In the corridor setting, the highest IoU values are obtained with 25 background scans ---$0.3812$ for individual sensors and $0.3878$ for combined--- while performance decreases with larger scan counts. In contrast, the best results in the intersection scenario are consistently achieved with only 10 background scans. Specifically, the IoU reaches $0.6972$ for individual 360-degree sensors, $0.8154$ for individual MEMS sensors, $0.6784$ for combined 360-degree sensors, $0.8073$ for combined MEMS sensors, and $0.7045$ when all sensors are used together. These results indicate that an excessive number of background scans can introduce variability or noise that negatively impacts segmentation accuracy, particularly in simpler and more constrained scenes such as the corridor.

{
An analysis of precision and recall metrics reveals further insights into the algorithm’s behavior across scenarios. In
\unskip\parfillskip 0pt \par}

\begin{center}
    \resizebox{0.46\textwidth}{!}{
    \centering
    \begin{tabular}{l|l|l|l|cccc}
        \multicolumn{8}{c}{\textbf{Foreground-Level Metrics}} \\
        \specialrule{1.3pt}{0pt}{0pt}
        \multicolumn{4}{l|}{Background files} & 10 & 25 & 50 & 100 \\
        \specialrule{1.3pt}{0pt}{0pt}
        
        \multirow{8}{*}{\rotatebox{90}{Corridor}} 
            & \multicolumn{2}{c|}{\multirow{4}{*}{\rotatebox{90}{\scriptsize Individual}}} 
                                        & Precision & $0.3141$ & $0.4681$ & $0.2005$ & $0.1273$  \\
            & \multicolumn{2}{c|}{}     & Recall    & $0.6913$ & $\mathbf{0.6724}$ & $0.6635$ & $0.6599$  \\
            & \multicolumn{2}{c|}{}     & F1 Score  & $0.4319$ & $0.5520$ & $0.3080$ & $0.2134$  \\
            & \multicolumn{2}{c|}{}     & IoU       & $0.2754$ & $0.3812$ & $0.1820$ & $0.1194$  \\
        \cline{2-8}
            & \multicolumn{2}{c|}{\multirow{4}{*}{\rotatebox{90}{\scriptsize Combined}}} 
                                        & Precision & $0.3155$ & $\mathbf{0.5074}$ & $0.2077$ & $0.1383$  \\
            & \multicolumn{2}{c|}{}     & Recall    & $0.6461$ & $0.6219$ & $0.6132$ & $0.6065$  \\
            & \multicolumn{2}{c|}{}     & F1 Score  & $0.4239$ & $\mathbf{0.5588}$ & $0.3103$ & $0.2252$  \\
            & \multicolumn{2}{c|}{}     & IoU       & $0.2690$ & $\mathbf{0.3878}$ & $0.1836$ & $0.1269$  \\
        \specialrule{1.3pt}{0pt}{0pt}

        \multirow{20}{*}{\rotatebox{90}{Intersection}} & \multirow{8}{*}{\rotatebox{90}{Individual}}  
            & \multirow{4}{*}{\rotatebox{90}{360}}   & Precision & $0.8039$ & $0.7438$ & $0.7350$ & $0.5419$  \\
        &   &                                           & Recall    & $0.8400$ & $0.8103$ & $0.7847$ & $0.7539$  \\
        &   &                                           & F1 Score  & $0.8216$ & $0.7756$ & $0.7591$ & $0.6306$  \\
        &   &                                           & IoU       & $0.6972$ & $0.6335$ & $0.6117$ & $0.4605$  \\
        \cline{3-8}
        &   & \multirow{4}{*}{\rotatebox{90}{MEMS}}  & Precision & $0.9102$ & $0.9452$ & $0.9520$ & $0.9547$  \\
        &   &                                           & Recall    & $\mathbf{0.8867}$ & $0.8549$ & $0.8330$ & $0.8027$  \\
        &   &                                           & F1 Score  & $\mathbf{0.8983}$ & $0.8978$ & $0.8885$ & $0.8721$  \\
        &   &                                           & IoU       & $\mathbf{0.8154}$ & $0.8145$ & $0.7994$ & $0.7732$  \\
        \cline{2-8}
        & \multirow{12}{*}{\rotatebox{90}{Combined}}
            & \multirow{4}{*}{\rotatebox{90}{360}}   & Precision & $0.7977$ & $0.7464$ & $0.7268$ & $0.5327$  \\
        &   &                                           & Recall    & $0.8194$ & $0.7287$ & $0.6843$ & $0.6379$  \\
        &   &                                           & F1 Score  & $0.8084$ & $0.7374$ & $0.7049$ & $0.5806$  \\
        &   &                                           & IoU       & $0.6784$ & $0.5841$ & $0.5443$ & $0.4091$  \\
        \cline{3-8}
        &   & \multirow{4}{*}{\rotatebox{90}{MEMS}}  & Precision & $0.9177$ & $0.9471$ & $0.9527$ & $\mathbf{0.9551}$  \\
        &   &                                           & Recall    & $0.8704$ & $0.8355$ & $0.8082$ & $0.7710$  \\
        &   &                                           & F1 Score  & $0.8934$ & $0.8879$ & $0.8745$ & $0.8532$  \\
        &   &                                           & IoU       & $0.8073$ & $0.7983$ & $0.7770$ & $0.7440$  \\
        \cline{3-8}
        &   & \multirow{4}{*}{\rotatebox{90}{All}}   & Precision & $0.8379$ & $0.8015$ & $0.7906$ & $0.6647$  \\
        &   &                                           & Recall    & $0.8157$ & $0.7124$ & $0.6612$ & $0.6100$  \\
        &   &                                           & F1 Score  & $0.8266$ & $0.7543$ & $0.7201$ & $0.6362$  \\
        &   &                                           & IoU       & $0.7045$ & $0.6056$ & $0.5626$ & $0.4665$  \\
        \specialrule{1.3pt}{0pt}{0pt}
    \end{tabular}}
    \captionof{table}{Foreground-level evaluation of the proposed method on corridor and intersection scenes from the RCooper~\cite{RCooper} Dataset using both single-LiDAR (Individual) and multi-LiDAR (Combined) configurations. For intersection scenes, Multiline 360\textdegree~and MEMS LiDAR sensors were analyzed. Best values for each metric are shown in bold.}
    \label{tab:iou}
    \vspace{-10pt}
\end{center}

the intersection setting, precision and recall values are generally well-balanced across all sensor configurations and background scan counts, indicating consistent detection performance. However, in the corridor scenario, a noticeable disparity is observed between these metrics. With 25 background scans, recall reaches $0.6724$ for individual sensors and $0.6219$ for combined sensors, while corresponding precision values are significantly lower at $0.4681$ and $0.5074$, respectively. This gap suggests that, in this scenario, the method tends to over-segment the foreground, resulting in many false positives. Consequently, this imbalance adversely affects both the F1 score and the IoU, reducing the overall effectiveness of segmentation in the corridor environment.

Moreover, the results highlight the advantage of using MEMS sensors in the intersection setting. Their denser, more detailed point clouds lead to higher performance compared to 360-degree multiline sensors. This advantage is also maintained when MEMS sensors are combined with 360-degree data, confirming their significant contribution to accurate background subtraction. Interestingly, in both scenarios, the method performs slightly better using sensors individually rather than in combination, suggesting sensor fusion may introduce inconsistencies that affect the algorithm’s effectiveness.

{In addition to evaluating performance at the point level, we also assess the quality of the background subtraction
\unskip\parfillskip 0pt \par}

\begin{center}
    \resizebox{0.46\textwidth}{!}{
    \centering
    \begin{tabular}{l|l|l|l|cccc}
        \multicolumn{8}{c}{\textbf{Object-Level Metrics}} \\
        \specialrule{1.3pt}{0pt}{0pt}
        \multicolumn{4}{l|}{Background files} & 10 & 25 & 50 & 100 \\
        \specialrule{1.3pt}{0pt}{0pt}
        
        \multirow{4}{*}{\rotatebox{90}{Corridor}} & \multicolumn{2}{c|}{\multirow{2}{*}{Individual}}
            & TPR           & $\mathbf{0.7973}$ & $0.7561$ & $0.7296$ & $0.7105$  \\
        &   \multicolumn{2}{c|}{}         & Completeness  & $0.6967$ & $0.6694$ & $0.6520$ & $0.6377$  \\
        \cline{2-8}
        & \multicolumn{2}{c|}{\multirow{2}{*}{Combined}}
            & TPR           & $0.8368$ & $0.7831$ & $0.7572$ & $0.7389$  \\
        &   \multicolumn{2}{c|}{}         & Completeness  & $\mathbf{0.7343}$ & $0.6997$ & $0.6806$ & $0.6631$  \\
        \specialrule{1.3pt}{0pt}{0pt}

        \multirow{10}{*}{\rotatebox{90}{Intersection}} & \multirow{4}{*}{\rotatebox{90}{\scriptsize Individual}} 
            & \multirow{2}{*}{\small 360}   & TPR           & $0.7659$ & $0.7526$ & $0.6613$ & $0.6437$  \\
        &   &                                              & Completeness  & $0.7011$ & $0.6798$ & $0.5998$ & $0.5801$  \\
        \cline{3-8}
        &   & \multirow{2}{*}{\small MEMS}   & TPR           & $\mathbf{0.8471}$ & $0.8242$ & $0.7991$ & $0.7793$  \\
        &   &                                              & Completeness  & $0.7876$ & $0.7711$ & $0.7517$ & $0.7210$  \\
        \cline{2-8}
        & \multirow{6}{*}{\rotatebox{90}{\scriptsize Combined}} 
            & \multirow{2}{*}{\small 360}   & TPR           & $0.7359$ & $0.7085$ & $0.5812$ & $0.5349$  \\
        &   &                                              & Completeness  & $0.6571$ & $0.6258$ & $0.5078$ & $0.4784$  \\
        \cline{3-8}
        &   & \multirow{2}{*}{\small MEMS}   & TPR           & $0.8178$ & $0.7868$ & $0.7544$ & $0.7290$  \\
        &   &                                              & Completeness  & $\mathbf{0.7499}$ & $0.7282$ & $0.7054$ & $0.6692$  \\
        \cline{3-8}
        &   & \multirow{2}{*}{\small All}    & TPR           & $0.7284$ & $0.6994$ & $0.5470$ & $0.5117$  \\
        &   &                                              & Completeness  & $0.6466$ & $0.6092$ & $0.4876$ & $0.4535$  \\
        \specialrule{1.3pt}{0pt}{0pt}
    \end{tabular}}
    \captionof{table}{Object-level evaluation of the proposed method on corridor and intersection scenes from the RCooper~\cite{RCooper} Dataset using both single-LiDAR (Individual) and multi-LiDAR (Combined) configurations. For intersection scenes, Multiline 360\textdegree~and MEMS LiDAR sensors were analyzed. Best values for each metric are shown in bold.}
    \label{tab:completeness}
\end{center}

at the object level. To this end, we use the true positive rate (TPR) to measure the proportion of correctly detected objects and propose a Completeness (Equation~\ref{eq:completeness}) metric to measure the proportion of detected points for each object. An object is considered correctly detected (true positive) if the percentage of detected points exceeds a predefined threshold. For our experiments, we use a threshold of $0.5$. The results for these metrics are presented in Table~\ref{tab:completeness}


\begin{equation} \text{Completeness} = \frac{1}{n}\sum_{i}^{n} \frac{TP_{pt}^{(i)}}{TP_{pt}^{(i)} + FN_{pt}^{(i)}} \label{eq:completeness} \end{equation}


\begin{figure*}[!ht]
    \centering
    \includegraphics[width=\textwidth]{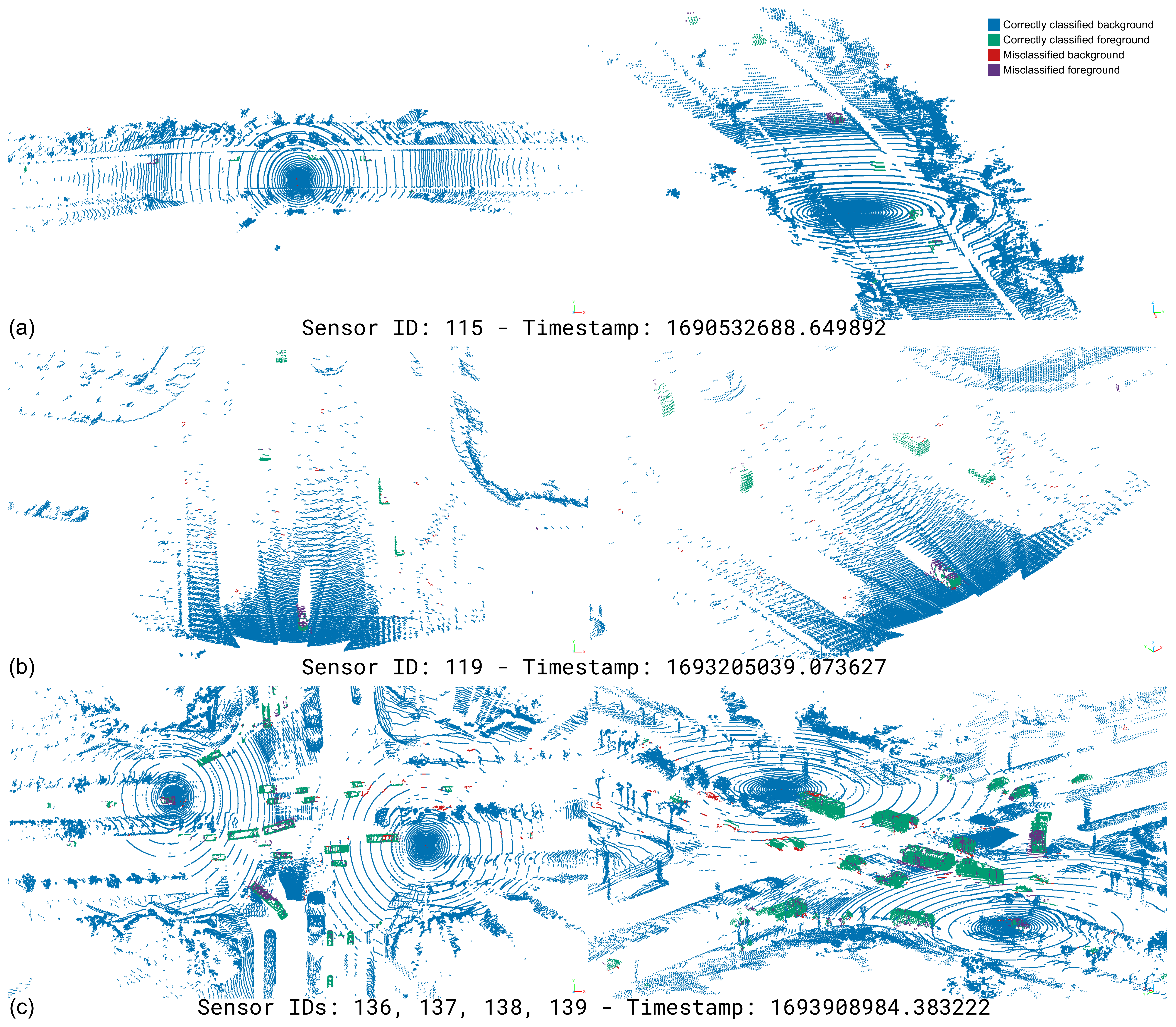}
    \caption{Qualitative results on the RCooper dataset~\cite{RCooper} with different sensor setups. (a) Corridor with a single multiline 360\textdegree{} LiDAR. (b) Intersection with a single MEMS LiDAR. (c) Intersection with multiple LiDARs ($2×$ 360\textdegree{} $+$ $2×$ MEMS). Blue: background (TP), green: foreground (TP), red: background (FN), purple: foreground (FN). Top view (left) and oblique view (right).}
    \label{fig:results}
\end{figure*}

The object-level evaluation confirms the effectiveness of the proposed method and complements the insights obtained from the point-level analysis. In the corridor scenario, the best results are obtained with 10 background scans, achieving a TPR of $0.7973$ and a Completeness of $0.6967$ for individual sensors, and slightly higher values of $0.8368$ and $0.7343$, respectively, when using combined sensors. In the intersection scenario, the best performance is also observed with 10 background scans across all sensor configurations. Specifically, individual 360-degree sensors achieve a TPR of $0.7659$ and Completeness of $0.7011$, while individual MEMS sensors reach the highest scores with a TPR of $0.8471$ and Completeness of $0.7876$. When using combined sensors, the 360-degree setup yields a TPR of $0.7359$ and Completeness of $0.6571$, MEMS sensors obtain $0.8178$ and $0.7499$, and the configuration including all sensors results in a TPR of $0.7284$ and Completeness of $0.6466$. These results reinforce previous observations that a moderate number of background scans leads to better performance, while increasing this number tends to degrade both metrics.

Compared to the point-level evaluation, the object-level metrics are more stable across different configurations and background scan counts. The differences between sensor types and scene setups are less pronounced, indicating that once an object is detected, it is generally detected in a consistent and complete manner regardless of the setup. This suggests that the proposed method, despite some variability at the point level, maintains reliable object-level performance.

We also provide qualitative results in both scenarios with different sensor configurations, as shown in Figure~\ref{fig:results}. In all cases, the proposed method successfully detects points belonging to all foreground objects, regardless of the sensor setup. The method performs consistently well with both individual and multi-sensor configurations. Furthermore, it introduces minimal background noise, demonstrating its robustness and effectiveness in complex urban environments.

\subsection{Comparative Evaluation}

To evaluate the contribution of our method, we compared the previously reported metrics with those obtained by a recent state-of-the-art background subtraction approach~\cite{22LuyangWang}. The results of this comparison are presented in Table~\ref{tab:compare}. Unlike our method, which is designed to be sensor-agnostic and flexible across multiple LiDAR configurations, the reference method is specifically tailored for a single rotating LiDAR sensor. Therefore, the comparison is restricted to set-ups involving a single rotative LiDAR data.

{
Initially, we evaluated the reference method using the same number of background scans as in our experiments. However, in contrast to our approach, the performance of the reference method improves as the number of background
\unskip\parfillskip 0pt \par}

\begin{center}
    \centering
    \resizebox{0.36\textwidth}{!}{
    \begin{tabular}{p{0.2cm}|p{1.7cm}|>{\centering\arraybackslash}p{1.4cm}>{\centering\arraybackslash}p{1.4cm}}
        \multicolumn{4}{c}{\textbf{Individual 360 LiDAR - Metrics Comparison}} \\
        \specialrule{1.3pt}{0pt}{0pt}
        \multicolumn{2}{c|}{} & Ref~\cite{22LuyangWang} & Ours \\
        \specialrule{1.3pt}{0pt}{0pt}
        \multirow{6}{*}{\rotatebox{90}{Corridor}} & \small Precision & $0.3800$ & $\mathbf{0.4681}$ \\
                                                   & \small Recall & $\mathbf{0.7241}$ & $0.6724$ \\
                                                   & \small F1 Score & $0.4985$ & $\mathbf{0.5520}$ \\
                                                   & \small IoU & $0.3320$ & $\mathbf{0.3812}$ \\
        \cdashline{2-4}
                                                   & \small TPR & $0.6450$ & $\mathbf{0.7561}$ \\
                                                   & \small Completeness & $0.5991$ & $\mathbf{0.6694}$ \\
        \hline
        \multirow{6}{*}{\rotatebox{90}{Intersection}} & \small Precision & $0.3800$ & $\mathbf{0.8039}$ \\
                                                  & \small Recall & $0.7241$ & $\mathbf{0.8400}$ \\
                                                  & \small F1 Score & $0.4985$ & $\mathbf{0.8216}$ \\
                                                  & \small IoU & $0.3320$ & $\mathbf{0.6972}$ \\
        \cdashline{2-4}
                                                  & \small TPR & $0.6450$ & $\mathbf{0.7659}$ \\
                                                  & \small Completeness & $0.5991$ & $\mathbf{0.7011}$ \\
        \specialrule{1.3pt}{0pt}{0pt}
    \end{tabular}}
    \captionof{table}{Performance comparison of the proposed method and the reference state-of-the-art approach~\cite{22LuyangWang} using a single Multiline 360° LiDAR on the RCooper dataset~\cite{RCooper}. Best values for each metric shown in bold.}
    \label{tab:compare}
\end{center}

scans increases. To ensure a fair comparison under optimal conditions for both methods, we used the maximum number of background scans available in the training split for the reference method, totaling 400 scans, and maintained the same hyperparameter settings as specified in its original publication.

The comparison demonstrates that our method consistently outperforms the reference approach~\cite{22LuyangWang} in nearly all evaluated metrics. In both the corridor and intersection scenarios, our method achieves significantly higher Intersection over Union (IoU) values, with $0.3812$ and $0.6972$ respectively, compared to $0.3320$ in both cases for the baseline. These improvements reflect a more accurate separation of foreground and background regions. Additionally, our approach obtains better scores in precision, F1 Score, TPR, and Completeness, while the reference method only shows a slight advantage in recall within the corridor scenario. Despite the reference method having access to a considerably larger set of background scans, our method not only remains competitive but clearly outperforms it, emphasizing the efficiency and generalization capability of our approach with minimal background data.

\begin{table*}
    \centering
    \resizebox{\textwidth}{!}{
    \begin{tabular}{l|l|l|l|ccccccccc}
        \multicolumn{13}{c}{\textbf{Metrics Per Class}} \\
        \specialrule{1.3pt}{0pt}{0pt}
        \multicolumn{4}{l|}{Class} & car & pedestrian & truck & bus & bicycle & motorcycle & tricycle & construction & \makecell{huge \\ vehicle} \\
        \specialrule{1.3pt}{0pt}{0pt}
        
        \multirow{6}{*}{\rotatebox{90}{Corridor}} 
            & \multicolumn{2}{c|}{\multirow{3}{*}{\rotatebox{90}{\tiny Individual}}} 
                                    & Recall & $\mathbf{0.6093}$ & $\mathbf{0.6636}$ & $\mathbf{0.7721}$ & $\mathbf{0.7300}$ & $\mathbf{0.6330}$ & $0.6732$ & $0.7241$ & $\mathbf{0.7560}$ & - \\
            \cdashline{4-13}
            & \multicolumn{2}{c|}{} & TPR & $0.7587$ & $0.7139$ & $\mathbf{0.9186}$ & $\mathbf{0.8962}$ & $0.6096$ & $0.7022$ & $0.8548$ & $\mathbf{0.9020}$ & - \\
            & \multicolumn{2}{c|}{} & Completeness & $0.6739$ & $0.6477$ & $\mathbf{0.7931}$ & $\mathbf{0.7534}$ & $0.5682$ & $0.6331$ & $0.6667$ & $\mathbf{0.7452}$ & - \\
        \cline{2-13} 
            & \multicolumn{2}{c|}{\multirow{3}{*}{\rotatebox{90}{\tiny Combined}}} 
                                    & Recall & $0.5483$ & $0.6499$ & $0.6937$ & $0.6964$ & $0.6069$ & $\mathbf{0.6687}$ & $\mathbf{0.7258}$ & $0.7126$ & - \\
            \cdashline{4-13}
            & \multicolumn{2}{c|}{} & TPR & $\mathbf{0.7616}$ & $\mathbf{0.7511}$ & $0.8770$ & $0.8799$ & $\mathbf{0.7027}$ & $\mathbf{0.7975}$ & $\mathbf{0.9612}$ & $0.7971$ & - \\
            & \multicolumn{2}{c|}{} & Completeness & $\mathbf{0.6850}$ & $\mathbf{0.6777}$ & $0.7569$ & $0.7454$ & $\mathbf{0.6648}$ & $\mathbf{0.7237}$ & $\mathbf{0.7478}$ & $0.7045$ & - \\
        \specialrule{1.3pt}{0pt}{0pt}

        \multirow{15}{*}{\rotatebox{90}{Intersection}} & \multirow{6}{*}{\rotatebox{90}{Individual}} 
            & \multirow{3}{*}{\rotatebox{90}{360}} & Recall & $0.7852$ & $0.4184$ & $0.8294$ & $0.8781$ & $\mathbf{0.9870}$ & $0.7903$ & $\mathbf{0.9591}$ & $\mathbf{0.9827}$ & $0.9314$ \\
        \cdashline{4-13}
        & & & TPR & $0.7865$ & $0.4888$ & $0.5826$ & $0.9712$ & $0.9167$ & $0.9073$ & $0.9302$ & $0.8333$ & $0.9608$ \\
        & & & Completeness & $0.7137$ & $0.4181$ & $0.5439$ & $0.8924$ & $0.9167$ & $\mathbf{0.8462}$ & $0.8830$ & $0.8121$ & $0.9030$ \\
        \cline{3-13} 
        & & \multirow{3}{*}{\rotatebox{90}{MEMS}} & Recall & $\mathbf{0.8273}$ & $0.1946$ & $\mathbf{0.8680}$ & $\mathbf{0.9781}$ & $0.5000$ & $\mathbf{0.8342}$ & $0.3939$ & $0.7018$ & $\mathbf{0.9341}$ \\
        \cdashline{4-13}
        & & & TPR & $\mathbf{0.8652}$ & $0.1911$ & $\mathbf{0.8779}$ & $0.9815$ & $0.4000$ & $0.7267$ & $0.2857$ & $0.8182$ & $0.9776$ \\
        & & & Completeness & $\mathbf{0.8027}$ & $0.1867$ & $\mathbf{0.7875}$ & $0.9446$ & $0.3333$ & $0.7090$ & $0.2857$ & $0.6564$ & $0.9189$ \\
        \cline{2-13} 
        & \multirow{9}{*}{\rotatebox{90}{Combined}} 
            & \multirow{3}{*}{\rotatebox{90}{360}} & Recall & $0.7733$ & $\mathbf{0.4184}$ & $0.8118$ & $0.8477$ & $\mathbf{1.0000}$ & $0.7694$ & $0.9602$ & $\mathbf{0.9827}$ & $0.8978$ \\
        \cdashline{4-13}
        & & & TPR & $0.7791$ & $\mathbf{0.6414}$ & $0.4512$ & $0.9858$ & $\mathbf{1.0000}$ & $0.9527$ & $\mathbf{1.0000}$ & $0.8333$ & $\mathbf{1.0000}$ \\
        & & & Completeness & $0.6918$ & $\mathbf{0.5473}$ & $0.4124$ & $0.8672$ & $\mathbf{1.0000}$ & $\mathbf{0.8509}$ & $0.9465$ & $0.8121$ & $0.9250$ \\
        \cline{3-13} 
        & & \multirow{3}{*}{\rotatebox{90}{MEMS}} & Recall & $0.8032$ & $0.1946$ & $0.8464$ & $0.9690$ & $0.5000$ & $0.8134$ & $0.3939$ & $0.7018$ & $0.9297$ \\
        \cdashline{4-13}
        & & & TPR & $0.8388$ & $0.1911$ & $0.8734$ & $\mathbf{0.9929}$ & $0.4000$ & $0.6866$ & $0.2857$ & $0.8182$ & $0.9967$ \\
        & & & Completeness & $0.7690$ & $0.1867$ & $0.7609$ & $\mathbf{0.9463}$ & $0.3333$ & $0.6688$ & $0.2857$ & $0.6564$ & $\mathbf{0.9337}$ \\
        \cline{3-13} 
        & & \multirow{3}{*}{\rotatebox{90}{All}} & Recall & $0.7581$ & $0.4103$ & $0.8110$ & $0.8642$ & $0.9960$ & $0.7707$ & $0.9600$ & $0.9508$ & $0.8829$ \\
        \cdashline{4-13}
        & & & TPR & $0.7735$ & $0.4990$ & $0.4504$ & $\mathbf{1.0000}$ & $\mathbf{1.0000}$ & $\mathbf{0.9529}$ & $\mathbf{1.0000}$ & $\mathbf{1.0000}$ & $\mathbf{1.0000}$ \\
        & & & Completeness & $0.6819$ & $0.4842$ & $0.4035$ & $0.8809$ & $0.9974$ & $0.8479$ & $\mathbf{0.9468}$ & $\mathbf{0.8047}$ & $0.9165$ \\
        \specialrule{1.3pt}{0pt}{0pt}
    \end{tabular}}
    \caption{Evaluation of the proposed method per class on corridor and intersection scenes from the RCooper dataset~\cite{RCooper}. The corridor evaluation was conducted using 10 background files, while the intersection evaluation used 25. Both analyses were performed with single-LiDAR and multi-LiDAR configurations. Best values for each metric are shown in bold.}
    \label{tab:classes}
    \vspace{-8pt}
\end{table*}

\begin{table*}
    \centering
    \resizebox{\textwidth}{!}{
    \begin{tabular}{l|l|l|cccc:c}
        \specialrule{1.3pt}{0pt}{0pt}
        \multicolumn{3}{l|}{} & Voxelization $O(N)$ & Point Count $O(M)$ & Background Filter $O(N)$ & ROR $O(K^2)$ & Total $O(N^2)$\\
        \specialrule{1.3pt}{0pt}{0pt}
        \multirow{4}{*}{\rotatebox{90}{Individual}} & \multirow{2}{*}{\rotatebox{90}{\tiny 360}}
        & Mean Measurements & $N = 88703.54$ & $M = 52139.53$ & $N = 88703.54$ & $K = 1238.98$ & $N = 88703.54$ \\
        & & Real time (ms) & $24.0648$ & $183.3969$ & $358.9785$ & $9.0122$ & $575.4525$ \\
        \cline{2-8}
        & \multirow{2}{*}{\rotatebox{90}{\tiny MEMS}}
        & Mean Measurements & $N = 40804.59$ & $M = 35725.63$ & $N = 40804.59$ & $K = 562.43$ & $N = 40804.59$ \\
        & & Real time (ms) & $11.4072$ & $115.8550$ & $164.6908$ & $6.2697$ & $298.2228$ \\
        \hline
        \multirow{6}{*}{\rotatebox{90}{Combined}} & \multirow{2}{*}{\rotatebox{90}{\tiny 360}}
        & Mean Measurements & $N = 177407.08$ & $M = 104170.34$ & $N = 177407.08$ & $K = 2466.53$ & $N = 177407.08$ \\
        & & Real time (ms) & $49.8069$ & $390.9233$ & $739.0308$ & $17.6002$ & $1197.3615$ \\
        \cline{2-8}
        & \multirow{2}{*}{\rotatebox{90}{\tiny MEMS}}
        & Mean Measurements & $N = 81609.19$ & $M = 71331.88$ & $N = 81609.19$ & $K = 1040.06$ & $N = 81609.19$ \\
        & & Real time (ms) & $23.3369$ & $242.2732$ & $344.3224$ & $10.7150$ & $620.6476$ \\
        \cline{2-8}
        & \multirow{2}{*}{\rotatebox{90}{\tiny All}}
        & Mean Measurements & $N = 259016.26$ & $M = 175031.24$ & $N = 259016.26$ & $K = 254096.97$ & $N = 259016.26$ \\
        & & Real time (ms) & $80.1094$ & $407.7107$ & $611.1274$ & $1483.3041$ & $2582.2519$ \\
        \specialrule{1.3pt}{0pt}{0pt}
    \end{tabular}}
    \caption{Average execution times (ms) and mean input sizes for each component of the proposed method, evaluated on a Jetson Nano 2GB Developer Kit. Results are reported separately for individual and combined LiDAR configurations, including both Multiline 360° and MEMS LiDAR. All measurements correspond to one hour of execution on the intersection split validation set.}
    \label{tab:time}
    \vspace{-10pt}
\end{table*}

\subsection{Per-Class Evaluation}

Finally, Table~\ref{tab:classes} presents the evaluation metrics per class. These tables report only Recall, True Positive Rate (TPR), and Completeness, as the remaining metrics require the number of false positives, which cannot be computed in our case. This limitation arises because our method does not perform semantic classification of points; it solely distinguishes foreground from background, making it impossible to determine false positives at the class level. Although this is not a complete evaluation, these results still provide valuable insight into which object categories pose more challenges for our method.

The per-class evaluation shows that, overall, the proposed method handles most object categories effectively. In the majority of cases, the method achieves high recall, TPR, and Completeness values—often exceeding $0.8$ and, in several instances, approaching values close to $1$ This consistent performance across diverse object types highlights the robustness of the algorithm in detecting and capturing the full extent of most foreground instances. Notably, classes such as \textit{car}, \textit{truck}, \textit{bus}, \textit{construction}, and \textit{huge vehicle} exhibit particularly strong results in both corridor and intersection scenarios, regardless of the sensor configuration.

However, a few classes present more challenges. Specifically, \textit{pedestrian}, \textit{bicycle}, \textit{motorcycle}, and \textit{tricycle} tend to yield lower scores across all three metrics. These classes typically correspond to smaller, thinner, or more dynamic objects, which are inherently more difficult to capture with high precision in LiDAR-based segmentation.

\subsection{Time Performance}

We evaluated the time performance of our method using a \textit{Jetson Nano 2GB Developer Kit}. This device was chosen specifically because it is low-cost and not particularly powerful, allowing us to demonstrate how our approach performs on resource-constrained hardware. Testing on such a device is important, as large-scale deployment of roadside LiDAR background subtraction for city-wide monitoring would likely require multiple sensors and embedded systems, making cost and efficiency critical factors.

The method is implemented in C++ using the Point Cloud Library (PCL). We evaluated the time performance of the proposed method by measuring the execution time of each component as well as the complete pipeline. The components include: (i) voxelization of the input point cloud, with time complexity $O(N)$, where $N$ is the number of input points; (ii) point counting per voxel cell, $O(M)$, with $M$ being the number of points after voxelization; (iii) background filtering, $O(N)$; and (iv) Radius Outlier Removal (ROR), with complexity $O(K^2)$, where $K$ is the number of foreground points proposed by the background filtering. The overall theoretical complexity of the full algorithm is $O(N^2)$.

Average execution times for each component are reported in Table~\ref{tab:time}. Additionally, we report mean values for $N$, $M$, and $K$, although these were not recorded during time measurements. As in the method validation, we separated time performance measurements according to the type of LiDAR sensor used, including individual sensors and combinations of them, allowing us to assess performance across varying sensor characteristics and input conditions. Runtime was measured over a continuous one-hour inference session using the validation split of the intersection scenes.

As shown in the time performance experiments, the proposed method runs efficiently on the \textit{Jetson Nano 2GB}, with runtimes of 298 ms for MEMS and 575 ms for 360-degree sensors in single-sensor setups. MEMS sensors yield faster processing despite providing denser point clouds, due to their more limited sensing range, which results in fewer overall points per scan. Execution time scales consistently with the theoretical complexity analysis, increasing with the number of input points and foreground elements, particularly in multi-sensor configurations where runtimes reach $2.5$ seconds.

Although the method does not achieve real-time performance on the Jetson Nano, this platform was selected to demonstrate feasibility under strict hardware constraints. In this context, the results provide a strong baseline. The most time-consuming stages are point counting and background filtering. However, the ROR step can become a bottleneck in scenes with high object density due to its quadratic complexity concerning the number of foreground points.

Despite this, the method is suitable for practical deployments. A more capable computing unit could handle multiple LiDAR streams centrally, eliminating the need for one processor per sensor and supporting flexible, scalable city-wide monitoring solutions.

\section{Conclusions and Future Work}

In this work, we have introduced a fully interpretable and statistically grounded approach for background subtraction in roadside LiDAR systems. Our method effectively models the spatial and height distributions of the background using only a small number of background-only scans. The two-phase process—comprising GDG generation and subsequent background subtraction—enables the method to operate with high flexibility across different LiDAR types and deployment scenarios, including both rotating and non-rotating sensor architectures.

Through extensive experiments on the publicly available RCooper dataset~\cite{RCooper}, we demonstrated that our approach consistently outperforms existing state-of-the-art techniques in terms of precision, recall, F1 score, IoU, and object-level metrics such as TPR and Completeness. Notably, the method exhibits strong generalization across different sensor configurations and scene types, with MEMS-based sensors providing particularly high performance. Moreover, our solution maintains competitive accuracy with minimal background data, further supporting its applicability in real-world deployments.

In terms of computational efficiency, our implementation achieves acceptable performance on resource-constrained hardware such as the Jetson Nano 2GB, confirming the feasibility of large-scale, embedded deployments for infrastructure-based perception systems. Although real-time performance is not achieved on this low-end platform, the results establish a solid baseline for further optimization.

Future work will address two key areas. First, we will focus on optimizing the implementation to reduce processing time and move closer to real-time performance, which is critical for deployment in dynamic environments. Second, we plan to extend the system with object classification capabilities, allowing it to detect foreground elements and identify their semantic categories, such as vehicles or pedestrians.

\section*{Acknowledgment}
This work has received funding from the Basque Government under project CAPACITY of the program HAZITEK-2024.
This work was partially supported by grant GIU23/022, funded by the University of the Basque Country (UPV/EHU).

\bibliographystyle{unsrt}  
\bibliography{references}  

\end{multicols}
\end{document}